\documentclass[conference]{IEEEtran}
\IEEEoverridecommandlockouts

\usepackage{cite}
\usepackage{amsmath,amssymb,amsfonts}
\usepackage{algorithmic}
\usepackage{graphicx}
\usepackage{textcomp}
\usepackage{xcolor}

\usepackage{hyperref}       
\usepackage{url}            
\usepackage{booktabs}       
\usepackage{amsfonts}       
\usepackage{nicefrac}       
\usepackage{microtype}      

\usepackage{graphicx}
\usepackage{multirow}
\usepackage{amssymb}
\usepackage{bbding}
\usepackage{adjustbox}
\usepackage{colortbl}
\usepackage{amsmath}
\usepackage{enumitem}
\usepackage{threeparttable}

\def\BibTeX{{\rm B\kern-.05em{\sc i\kern-.025em b}\kern-.08em
    T\kern-.1667em\lower.7ex\hbox{E}\kern-.125emX}}
\begin{document}

\title{Alifuse: Aligning and Fusing Multimodal Medical Data for Computer-Aided Diagnosis\\
}

\author{\IEEEauthorblockN{Qiuhui Chen, and Yi Hong\textsuperscript{*}}
\IEEEauthorblockA{\textit{Department of Computer Science and Engineering, Shanghai Jiao Tong University
Shanghai 200240, China} \\
\textsuperscript{*}Corresponding author. Email: yi.hong@sjtu.edu.cn}}


\maketitle

\begin{abstract}
Medical data collected for diagnostic decisions are typically multimodal, providing comprehensive information on a subject. While computer-aided diagnosis systems can benefit from multimodal inputs, effectively fusing such data remains a challenging task and a key focus in medical research. In this paper, we propose a transformer-based framework, called Alifuse, for aligning and fusing multimodal medical data. Specifically, we convert medical images and both unstructured and structured clinical records into vision and language tokens, employing intramodal and intermodal attention mechanisms to learn unified representations of all imaging and non-imaging data for classification. Additionally, we integrate restoration modeling with contrastive learning frameworks, jointly learning the high-level semantic alignment between images and texts and the low-level understanding of one modality with the help of another. We apply Alifuse to classify Alzheimer’s disease, achieving state-of-the-art performance on five public datasets and outperforming eight baselines. The source code is available at \href{https://github.com/Qybc/Alifuse}{https://github.com/Qybc/Alifuse}.
\end{abstract}

\vspace{-0.2em}
\section{Introduction}

Recent advances in deep neural networks have significantly influenced the field of medical image analysis, demonstrating notable successes across various medical research domains, particularly in computer-aided diagnosis (CAD)~\cite{chan2020computer}. Medical images are pivotal components in clinical diagnosis
. 
however, diagnostic data typically manifests in a multimodal nature, encompassing medical images alongside non-imaging data like patient demographics and laboratory findings. For instance, diagnosing Alzheimer’s Disease (AD) requires clinicians to consider medical brain scans, demographics, laboratory test results, and other clinical observations to make accurate diagnostic decisions. The automatic integration of these diverse data types is crucial for effective computer-aided diagnosis.

A common approach in recent research involves extracting features from imaging and non-imaging data separately, followed by their feature fusion for diagnostic classification~\cite{
chen2023medblip}. Popular models in large vision language pre-training (VLP), such as CLIP~\cite{radford2021learning},
have shown promising performance in various computer vision downstream applications, including classification
~\cite{bao2022vlmo} 
and segmentation~\cite{xu2021simple}. These VLP models learn multimodal representations from large image and text datasets by aligning their features into a common space. Researchers have adopted these techniques to generate medical reports~\cite{yan2022clinical}, where the texts describe medical images. In recent work, MedCLIP~\cite{wang2022medclip} integrates imaging data into the textual space created by non-imaging data, using this alignment with a large language model to facilitate question-answering. 
Despite these advancements, integrating diverse data types for CAD remains challenging due to the distinct nature of imaging and non-imaging data, which often exist in disparate high-dimensional spaces. Additionally, aligning these multimodal features into a unified space for effective fusion is complex. 

\begin{figure*}[t]
  \centering
  \includegraphics[width=0.9\linewidth]{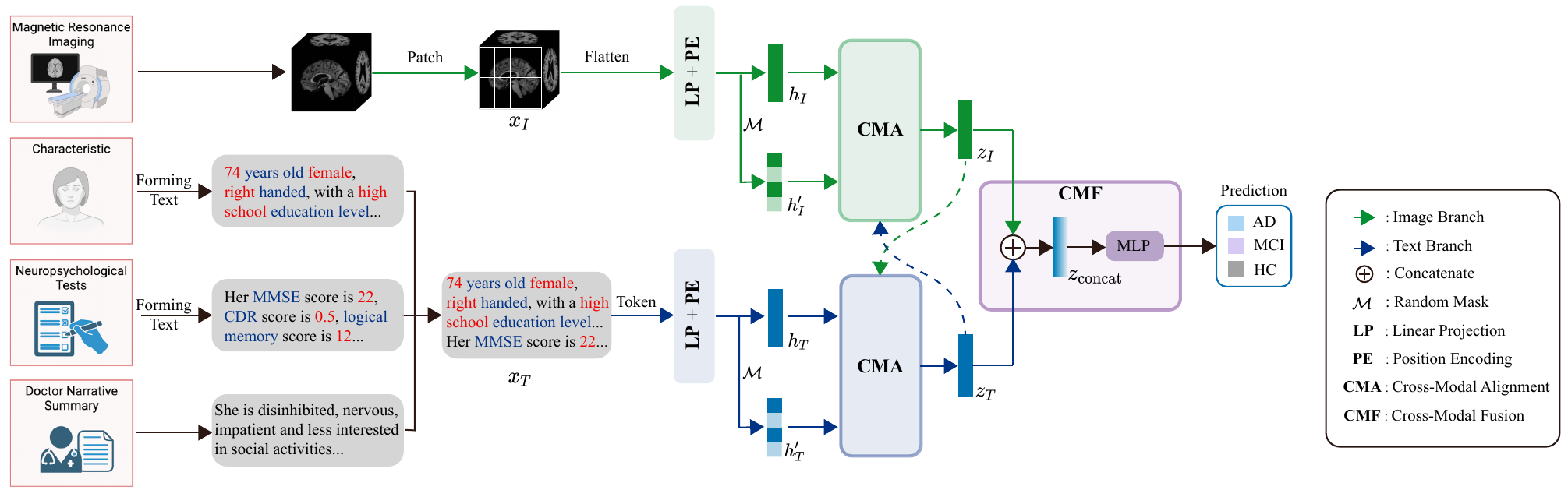}
  \caption{The architecture overview of our proposed model \textbf{Alifuse}, a CAD system designed for medical diagnosis using electronic health records through multimodal alignment and fusion. The cross-modal alignment (CMA) is a key component of Alifuse, which is illustrated in Fig.~\ref{fig:cma}.
  }
  \label{fig:overview}
\end{figure*}

In this paper, we address these challenges by proposing a CAD model called Alifuse
. Alifuse offers a comprehensive framework for incorporating diverse data sources in clinical decision-making, as illustrated in Fig.~\ref{fig:overview}. For imaging data, such as MRI scans, we utilize a 3D vision Transformer~\cite{dosovitskiy2020image} to extract volumetric image features. For non-imaging data from electronic health records, including demographics, laboratory test results, and doctor's narrative summaries, we convert this information into text form to enhance contextual informativeness and extract textual features using a language model like BERT~\cite{devlin2018bert}. The alignment of imaging and non-imaging features is achieved through image-text contrastive learning; simultaneously, a fusion process employs cross-attention on the aligned features. The contrastive-learning-based alignment focuses on high-level semantics, resulting in the loss of low-level detail understanding and the fine-grained correspondence between them. Therefore, we propose using masked image and text reconstruction to improve the mutual understanding between images and texts. This integrated approach ensures balanced consideration of both imaging and non-imaging data, leading to improved diagnostic accuracy and comprehensive clinical decision-making support.

We utilize Alifuse to distinguish AD and mild cognitive impairment (MCI) subjects from normal controls (NC), leveraging both imaging and non-imaging data from electronic health records. 
Our dataset comprises 27.8K medical image volumes and additional clinical data sourced from five public multimodal AD datasets, including ADNI~\cite{petersen2010alzheimer}, NACC~\cite{beekly2007national}, OASIS~\cite{marcus2007open},
AIBL~\cite{ellis2009australian}, and MIRIAD~\cite{malone2013miriad}, for training and evaluating our Alifuse model. 
Also, we conduct a comparative analysis with eight baselines to showcase the effectiveness of our approach. Our contributions are summarized as follows:
\begin{itemize}
    \item We present Alifuse, a unified CAD framework designed to align and fuse multimodal medical inputs, integrating medical scans with non-imaging clinical records to support well-informed diagnostic decisions.  

    \item Alifuse implements a new collaborative learning paradigm that combines semantic alignment and restoration learning into unified representations for aligning and fusing multimodal medical information.

    \item Our model, trained and evaluated on a large-scale dataset for AD study, achieves state-of-the-art (SOTA) performance in AD classification. Alifuse accepts raw images without complex pre-processing and efficiently handles missing non-imaging data, enhancing its practical usability. Additionally, it has the potential to accommodate additional modalities and explore diseases beyond AD.





\end{itemize}

\begin{figure}[t]
  \centering
  \includegraphics[width=0.90\linewidth]{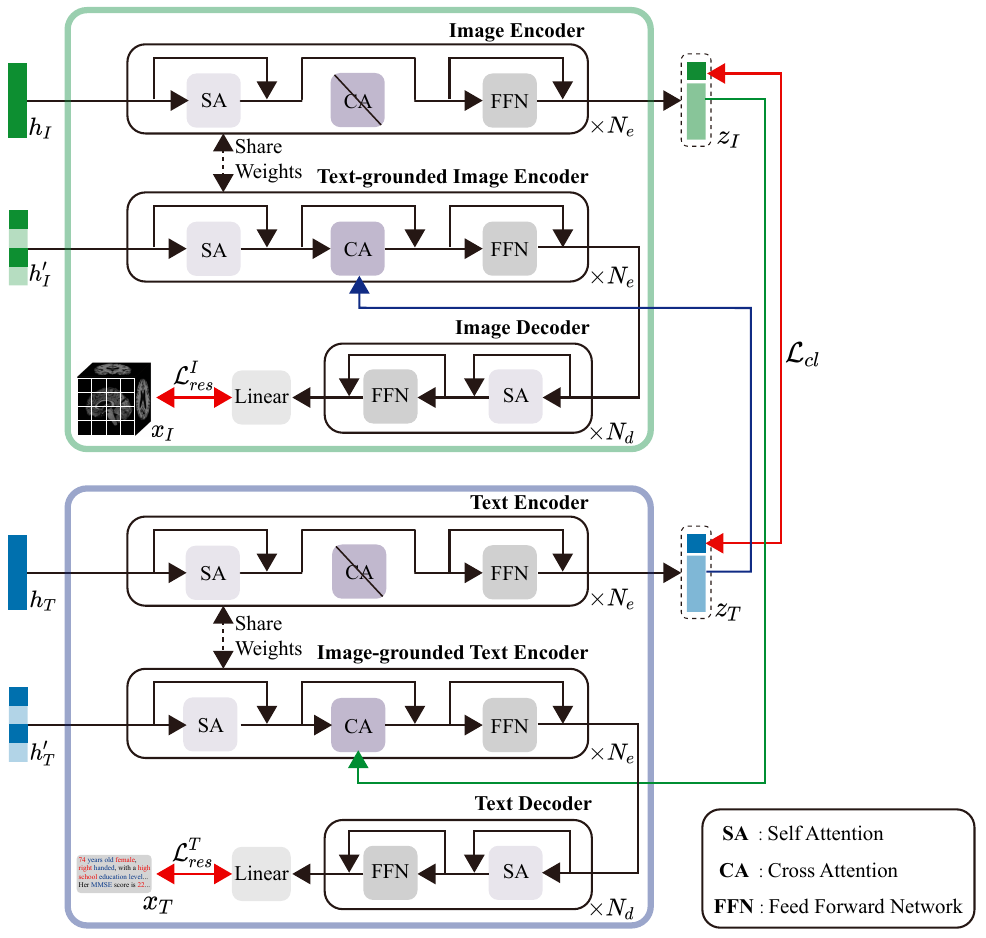}
  \caption{The architecture of the cross-modal alignment (CMA) module.
  }
  \label{fig:cma}
\end{figure}

\section{Methodology}
Figure~\ref{fig:overview} illustrates the comprehensive framework of our proposed model, Alifuse, for AD diagnosis. 

\textbf{Multimodal Data Collection.}
This study collects multimodal data from five public AD datasets.
The various modalities include imaging data, such as structural MRIs, unstructured non-imaging data (e.g., doctor's narrative summary and diagnosis), and structured non-imaging data (e.g., demographics and laboratory test results). This combination of clinical records and imaging data enables the model to learn about the AD characteristics of a subject comprehensively. 
We set the maximum text length to 40 words to process the unstructured doctor's narrative summaries considering the length of most data. If a summary exceeds 40 words, the tails will be truncated; otherwise, zero padding is applied to meet the length requirement. 
For structured clinical information, like MMSE (Mini-Mental State Examination) scores, CDR (Clinical Dementia Rating), and logical memory scores, we argue that the numerical values and categories are more meaningful in textual context. Therefore, we create a text representation for each data sample using the information from the raw data tables. For instance, if a patient has an MMSE score of 29, the corresponding text is “The MMSE score is 29”. As shown in Fig.~\ref{fig:overview}, all non-imaging data are converted into text representations to serve as inputs for the model.

\textbf{Framework Overview.}
As depicted in Fig.~\ref{fig:overview}, Alifuse processes a 3D image volume and corresponding text inputs by splitting them into image patches $x_I$ and text tokens $x_T$. 
Following a low-dimensional linear projection (LP) of the flattened image patches and text tokens, a learnable position embedding (PE) is incorporated into these representations, producing vector representations $h_I$ and $h_T$ for the image and text, respectively.
A Cross-Modal Alignment (CMA) module is then designed to extract and align their features, enhancing the following-up fusion process. Finally, the aligned image feature $z_I$ and text feature $z_T$ are fused through a Cross-Modal Fusion (CMF) module for diagnostic prediction.

\textbf{Cross-Modal Alignment (CMA) Module.}
\label{sec:cma}
To achieve multimodal feature alignment, we focus on the agreement between multimodal features and their ability to reconstruct masked data within the context of another modality. Therefore, we propose a multimodal encoder-decoder mixture model that employs multi-task learning through contrastive and restorative learning techniques, as illustrated in Fig.~\ref{fig:cma}. This dual-focus design enhances feature alignment by optimizing feature cross-modal agreement and enabling cross-modality data reconstruction.
Specifically, the CMA module comprises two unimodal encoders, two cross-modal-grounded encoders, and two unimodal decoders. 

\subsubsection{Unimodal Encoder}
We utilize a unimodal encoder for each modality to manage image and text inputs.
The image encoder, depicted at the top of the green box in Fig.~\ref{fig:cma}, is a 3D vision transformer~\cite{dosovitskiy2020image}. It encodes image vector representation $h_I$ as an image embedding $z_I$, including an additional $[CLS]$ token to represent the global image feature. 
The text encoder employs the BERT transformer~\cite{devlin2018bert} to map the text representation $h_T$ to its feature representation $z_T$. Similarly, a $[CLS]$ token is appended to the beginning of the text feature to summarize the input sentence. Each unimodal encoder comprises a self-attention (SA) layer and a feed-forward network (FFN) in each transformer block.

\subsubsection{Cross-Modal-Grounded Encoder} 
Unlike the unimodal encoder, the text-grounded image encoder integrates the textual feature $z_T$ by inserting an additional cross-attention (CA) layer between the SA layer and FFN for each transformer block of the image encoder. The two image encoders share all weights except for the CA layers. Similarly, the image-grounded text encoder incorporates the image feature $z_I$ by adding a cross-attention (CA) layer and shares weights with the text encoder described above. Both cross-modal-grounded encoders utilize the masked inputs of one modality, i.e., $h'_I$ or $h'_T$, to enhance contextual understanding within the context of another modal feature. We extract and align the image and text features by jointly training the unimodal and cross-modal-grounded encoders and applying contractive learning to their global feature [CLS] tokens. 


\subsubsection{Unimodal Decoder}

The contractive learning approach described above facilitates efficient semantic feature extraction between images and text. To retain fine-grained understanding of unimodal data, we incorporate a unimodal decoder following the cross-modal-grounded text encoder. This lightweight decoder consists of eight transformer blocks with an embedding dimension of 768
. Each transformer block comprises a self-attention (SA) layer and a feed-forward network (FFN), followed by a image/text decoder linear projection layer to reconstruct the masked image patches $x_I$ or text tokens $x_T$. 

\textbf{Cross-Modal Fusion (CMF) Module.}
After alignment by CMA, the image feature representation $z_I$ and text feature representation $z_T$ are concatenated to form a combined feature vector $z_{\text{concat}}=[z_I,z_T]$. This concatenated feature is fed into a multi-layer perception (MLP) with a hidden layer activated by ReLU.
The MLP's output is subsequently processed via a softmax function
to predict the final disease category.

\textbf{Learning Objectives.}
The CMA module employs two types of loss functions for learning, i.e., contrastive learning loss and reconstruction loss for self-learning. Meanwhile, the CMF module uses a classification loss for supervised learning. 

\subsubsection{Contrastive Learning Loss}
We utilize the Image-Text Contrastive (ITC) loss to align the feature spaces $z_I$ and $z_T$, produced by the image encoder and text encoder. The ITC loss promotes similarity in representations for positive image-text pairs and discourages similarity for negative pairs~\cite{radford2021learning}.

Assume we denote $D = \{(x_I^i,x_T^i)\}_{i=1}^{n}$ as all pairs of image and text tokens in the training dataset, $f_{I(\theta)}$ and $f_{T(\phi)}$ as the image and text encoder, respectively. 
Then, the objective function of image-text contrastive learning is formulated as:
\begin{equation}
\begin{aligned}
& \mathcal{L}_{cl} = \mathbb{E}_{(X_I^b,X_T^b)\sim D} \\ 
& \Biggl[ -\sum_{i=1}^{b} \log \left(\frac{\exp (s(f_{I(\theta)}(x_I^i),f_{T(\phi)}(x_T^i)) / \tau)}{\sum_{k=1}^{b} \exp (s(f_{I(\theta)}(x_I^i),f_{T(\phi)}(x_T^k)) / \tau)}\right) \\
& -\sum_{i=1}^{b} \log \left(\frac{\exp (s(f_{I(\theta)}(x_I^i),f_{T(\phi)}(x_T^i)) / \tau)}{\sum_{k=1}^{b} \exp (s(f_{I(\theta)}(x_I^k),f_{T(\phi)}(x_T^i)) / \tau)}\right) 
\Biggr]
\end{aligned}
\end{equation}
where $(X_I^b,X_T^b)$ is a minibatch of $b$ image-text pairs drawn from $D$, $\tau$ is the temperature hyperparameter, and $s(\cdot, \cdot)$ is the cosine similarity between image and text embeddings.

\subsubsection{Reconstruction Loss}
Our reconstruction learning module is designed to enhance the above global semantic understanding by utilizing fine-grained visual and textual information. 
Each image-text pair requires a second forward pass through each cross-modal-grounded encoder.
As illustrated in Fig~\ref{fig:cma}, the visual component of CMA involves a reconstruction learning branch consisting of a text-grounded image encoder $f_{I(\theta)}'$ and an image decoder $g_{I(\psi)}$. The text-grounded image encoder $f_{I(\theta)}'$ is shared with the image encoder $f_{I(\theta)}$. 
Given the representation $h_I$ of the image input $x_I$, which is masked by a random mask $\mathcal{M}$, i.e., $h'_I = \mathcal{M}(h_I)$, $f_{I(\theta)}'$ and $g_{I(\psi)}$ aim to reconstruct the original image from the masked representation, i.e.,  
$f_{I(\theta)}', g_{I(\psi)}: h'_I \rightarrow x'_I$, formally,  $x_I' = g_{I(\psi)}(f_{I(\theta)}'(\mathcal{M}(h_I)))$. Specially,
$f_{I(\theta)}'$ and $g_{I(\psi)}$ are trained by minimizing the pixel-level distance between the original and reconstructed images:
\begin{equation}
\label{eq:image_res_loss}
    \mathcal{L}_{res}^I =  \mathbb {E}_{x_I} \; \mathcal{D}_I(x_I, x_I'),
\end{equation}
where $\mathcal{D}_I(x_I, x_I')$ presents the distance function that measures similarity between $x_I$ and $x_I'$, e.g., Mean Square Error (MSE), or L1 norm. We use MSE following the common setting~\cite{he2022masked}.

For the textual component of CMA, we employ analogous operations. We define the text-grounded image encoder as $f_{T(\phi)}'$ and the text decoder as $g_{T(\eta)}$. Both $f_{T(\phi)}'$ and $g_{T(\eta)}$ are trained by minimizing the token-level distance between the original text $x_T$ and the reconstructed text $x'_T$:
\begin{equation}
\label{eq:text_res_loss}
    \mathcal{L}_{res}^T = \mathbb {E}_{x_T} \; \mathcal{D}_T(x_T, x_T'),
\end{equation}
where $x_T' = g_{T(\eta)}(f_{T(\phi)}'(\mathcal{M}(h_T)))$ denotes the reconstructed one, $\mathcal{M}$ is a random mask, $h_T$ is the vector representation of $x_T$, and
$\mathcal{D}_T(x_T, x_T')$ is the distance function measuring text similarity, such as the commonly-used cross-entropy loss.

The CMF module utilizes ground-truth labels of each subject for supervision. Let the disease prediction for the $i$-th subject be denoted as $\hat{y}^i$. With the ground truth label $y^i$, the cross-entropy objective for disease prediction can be defined as follows:
\begin{equation}
    \mathcal{L}_{cls} = - \frac{1}{N} \sum_{i=1}^{N} y^i \text{log} (\hat{y}^i).
\end{equation}

During training, we jointly optimize four objectives: one contractive learning loss, two reconstruction losses, and one classification loss, which together comprise the total loss:
\begin{equation}
    \mathcal{L} = \lambda_{cl} \mathcal{L}_{cl} + \lambda_{res} (\mathcal{L}_{res}^I + \mathcal{L}_{res}^T) + \lambda_{cls} \mathcal{L}_{cls},
\end{equation}
where $\lambda_{cl}$, $\lambda_{res}$, and $\lambda_{cls}$ are constant factors that balance these terms. Each image-text pair necessitates two forward passes through the visual and text transformers, with each pass activating different loss functions for optimization.  

Through our unified training scheme, Alifuse learns a representation that preserves fine-grained details within the samples while achieving global feature alignment between images and text. 
Specifically, the formulation of $\mathcal{L}_{cl}$ encourages the model to capture high-level discriminative features; $\mathcal{L}_{res}$ compels the model to encode fine-grained information from the images and texts by focusing on pixel/token-level patterns. 
This results in descriptive feature embeddings that enhance the discrimination task. Finally, $\mathcal{L}_{cls}$ promotes task-relevant learning by capturing disease-diagnosis-informative features.

\vspace{-0.2em}

\section{Experiments}


\begin{table}
\scriptsize
\centering
\caption{Dataset statistics. $\cdot$ Demogr.: Demographics, e.g., age, gender, hand, etc., $\cdot$ LabRes.: Laboratory Test Results, $\cdot$ Dr.Sum.: Doctor Narrative Summary, $\cdot$ CLS: class label.}
\begin{tabular}{c|ccccc}
    \toprule
    Datasets & \#Img. & \#Demogr. & \#LabRes. & Dr.Sum. & \#CLS \\
    \midrule
    ADNI~\cite{petersen2010alzheimer} & 10387 & 3 & 3 & \checkmark & 3\\
    NACC~\cite{beekly2007national} & 15343 & 4 & 3 & $\times$ & 3 \\
    AIBL~\cite{ellis2009australian}  & 1002 & 2 & 3 & $\times$ & 3 \\
    OASIS2~\cite{marcus2007open} & 373 & 3 & 3 & $\times$ & 2 \\
    MARIAD~\cite{malone2013miriad}  & 708 & 2 & 2 & $\times$ & 2 \\
    
  \bottomrule
\end{tabular}
\label{tab:data}
    \parbox{\linewidth}{
        \footnotesize \textit{Note: The numbers in the right four columns represent the number of clinical data types, not the number of subjects or data items.}}
\end{table}

\subsection{Datasets and Metrics}
To evaluate our proposed method, we conduct experiments on five multimodal public datasets, ADNI~\cite{petersen2010alzheimer}, NACC~\cite{beekly2007national}, AIBL~\cite{ellis2009australian}, OASIS2~\cite{marcus2007open} and MARIAD~\cite{malone2013miriad}, all focused on the study of Alzheimer's Disease (AD). 
Our model is assessed for its ability to distinguish subjects with AD or mild cognitive impairment (MCI) from normal controls (NC). Detailed statistics of these datasets are presented in Table~\ref{tab:data}. In total, we utilize 27.8K image volumes, most of which include demographic information and laboratory test results. Notably, only the ADNI dataset contains doctors' narrative summaries. 
Due to the varying volume sizes and image spacing, we pad each image to form a cube shape and then scale it to a unified size of $128 \times 128 \times 128$ for input. 
The image intensity of each volume is normalized to the range [0, 1].

We split ADNI and NACC datasets on a subject-wise basis into training, validation, and test sets, with proportions of 70\%, 10\%, and 20\%, respectively, yielding approximately 20K samples for training. In addition to evaluating ADNI and NACC test sets, we perform zero-shot testing on the OASIS2, AIBL, and MIRIAD datasets. Our evaluation metrics include accuracy and the area under the ROC curve (AUC) for classification.

\subsection{Baseline Methods} 
We compare Alifuse with three groups of baseline methods: 

\subsubsection{Image-only group}
This group includes four recent methods commonly used in image classification, including 3D ResNet50~\cite{he2016deep}, MedicalNet~\cite{chen2019med3d}, 3D Vit~\cite{dosovitskiy2020image}, and M3T~\cite{jang2022m3t}. 

\subsubsection{Text-only group}
To the best of our knowledge, no current studies are using medical non-imaging data alone for diagnosis. Therefore, we select BERT~\cite{devlin2018bert} as our text-only baseline. 

\subsubsection{Multimodal group}
This group includes three baseline methods: GIT~\cite{wang2022git}, Preciver~\cite{jaegle2021perceiver}, and IRENE~\cite{zhou2023transformer}, which are recent transformer-based models that fuse multimodal information for classification. 

\subsection{Implementations details}
For pre-processing, each image is resized and cropped to $128 \times 128 \times 128$, with an image patch size of $16$. The maximum input text length is set to 70. Following a common setting in~\cite{he2022masked}, we set all hidden feature sizes to 768 and the number of heads in multi-head attention (MHA) to 12. The number of layers in the encoder and cross-modal grounded encoder (i.e., $N_e$) is set to 24, while the number of layers in the decoder (i.e., $N_d$) is set to 8. The mask ratio of the random mask $\mathcal{M}$ is 0.5. 
The distance function in Eq.~\ref{eq:image_res_loss} is the $L_2$ norm and in Eq.~\ref{eq:text_res_loss}, it is Cross Entropy (CE). The values for $\lambda_{cl}, \lambda_{res}, \lambda_{cls}$ are empirically set to 1, 1, and 1, respectively. Our model is trained on four NVIDIA RTX 3090 GPUs with a batch size of $24$, using the AdamW optimizer with a learning rate of 2e-5.


\begin{table*}[t]
\scriptsize
\centering
\caption{Comparison of our method with eight baselines on five public datasets. $\dagger$: zero-shot classification. 
}
  \begin{tabular}{l|l|cc|cc|cc|cc|cc}
    \toprule
    \multicolumn{2}{c|}{\multirow{2}*{Methods}} & \multicolumn{2}{c|}{ADNI-test$^1$} & \multicolumn{2}{c|}{NACC-test$^1$} & \multicolumn{2}{c|}{AIBL$^1${\quad$\dagger$}} & \multicolumn{2}{c|}{OASIS2$^2${\quad$\dagger$}} & \multicolumn{2}{c}{MIRIAD$^2${\quad$\dagger$}} \\
    \cline{3-12}
    \multicolumn{2}{c|}{~} & ACC & AUC & ACC & AUC & ACC & AUC & ACC & AUC & ACC & AUC \\
    \cline{1-12}
    \multirow{4}*{Image-only} & 3D ResNet50~\cite{he2016deep} & 65.22\% & 73.91\% & 68.18\% & 76.14\% & 30.29\% & 47.72\% & 63.64\% & 72.73\% & 62.50\% & 71.88\% \\
    & MedicalNet~\cite{chen2019med3d} & 69.57\% & 77.17\% & 71.67\% & 77.25\% & 39.64\% & 54.73\% & 65.83\% & 73.75\% & 73.68\% & 80.26\% \\
    & 3D ViT~\cite{dosovitskiy2020image} & 72.04\% & 79.03\% & 76.19\% & 82.14\% & 45.54\% & 59.15\% & 69.57\% & 77.17\% & 77.27\% & 82.95\% \\
    & M3T~\cite{jang2022m3t} & 74.77\% & 81.08\% & 75.47\% & 81.60\% & 59.83\% & 69.87\% & 69.09\% & 76.82\% & 75.21\% & 81.40\% \\
    \cline{1-12}
    Text-only & BERT~\cite{devlin2018bert} & 81.80\% & 86.35\% & 80.18\% & 85.14\% & 80.64\% & 85.48\% & 89.02\% & 91.76\% & 71.68\% & 78.76\% \\
    \cline{1-12}
    \multirow{4}*{Multimodal} & Perceiver~\cite{jaegle2021perceiver} & 73.91\% & 80.43\% & 72.73\% & 79.55\% & 74.48\% & 78.51\% & 82.57\% & 86.93\% & 72.73\% & 79.55\% \\
    & GIT~\cite{wang2022git} & 82.46\% & 86.84\% & 79.65\% & 84.73\% & 77.90\% & 82.14\% & 84.21\% & 88.16\% & 77.68\% & 83.26\% \\
    & Irene~\cite{zhou2023transformer} & 83.48\% & 87.61\% & 80.95\% & 85.71\% & 82.30\% & 86.73\% & 89.29\% & 91.06\% & 79.02\% & 84.27\%  \\
    & Alifuse (ours) & \textbf{87.03\%} & \textbf{90.28\%} & \textbf{83.01\%} & \textbf{87.26\%} & \textbf{88.26\%} & \textbf{91.20\%} & \textbf{92.81\%} & \textbf{94.36\%} & \textbf{85.09\%} & \textbf{88.82\%}  \\
    \bottomrule
\end{tabular}
\begin{tablenotes}
\footnotesize
\item $^1$ Three-label classification: NC, MCI, and AD/Dementia. 
$^2$ Binary classification: NC and Alzheimer's sufferers (including MCI and AD/Dementia).
\end{tablenotes}
\label{tab:overall}
\end{table*}

\begin{figure}[t]
  \centering
  \includegraphics[width=1.0\linewidth]{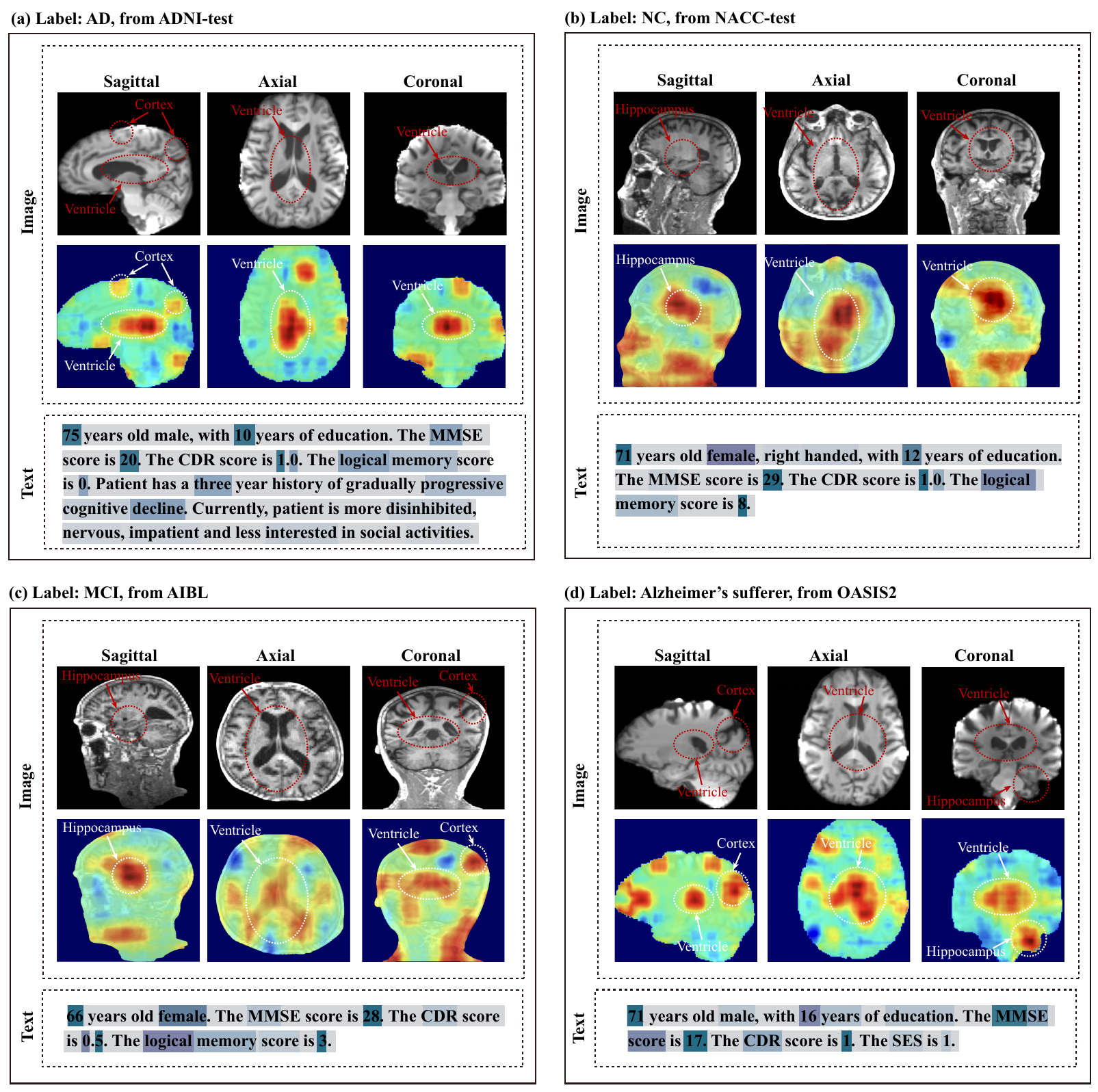}
  \caption{Visualization of attention maps using transformer interpretability techniques. 
  (a) ADNI test datasets, (b) NACC test dataset, (c) AIBL dataset, and (d) OASIS2 dataset. The image heatmap uses a jet colormap, where red indicates high activated values (close to one) and blue indicates low activated values (close to zero). A blue colormap is employed for the text heatmap, with darker blue representing higher attention levels. (Best viewed in color)}
  \label{fig:attention}
\end{figure}

\begin{figure}[htbp]
  \centering
  \includegraphics[width=1.0\linewidth]{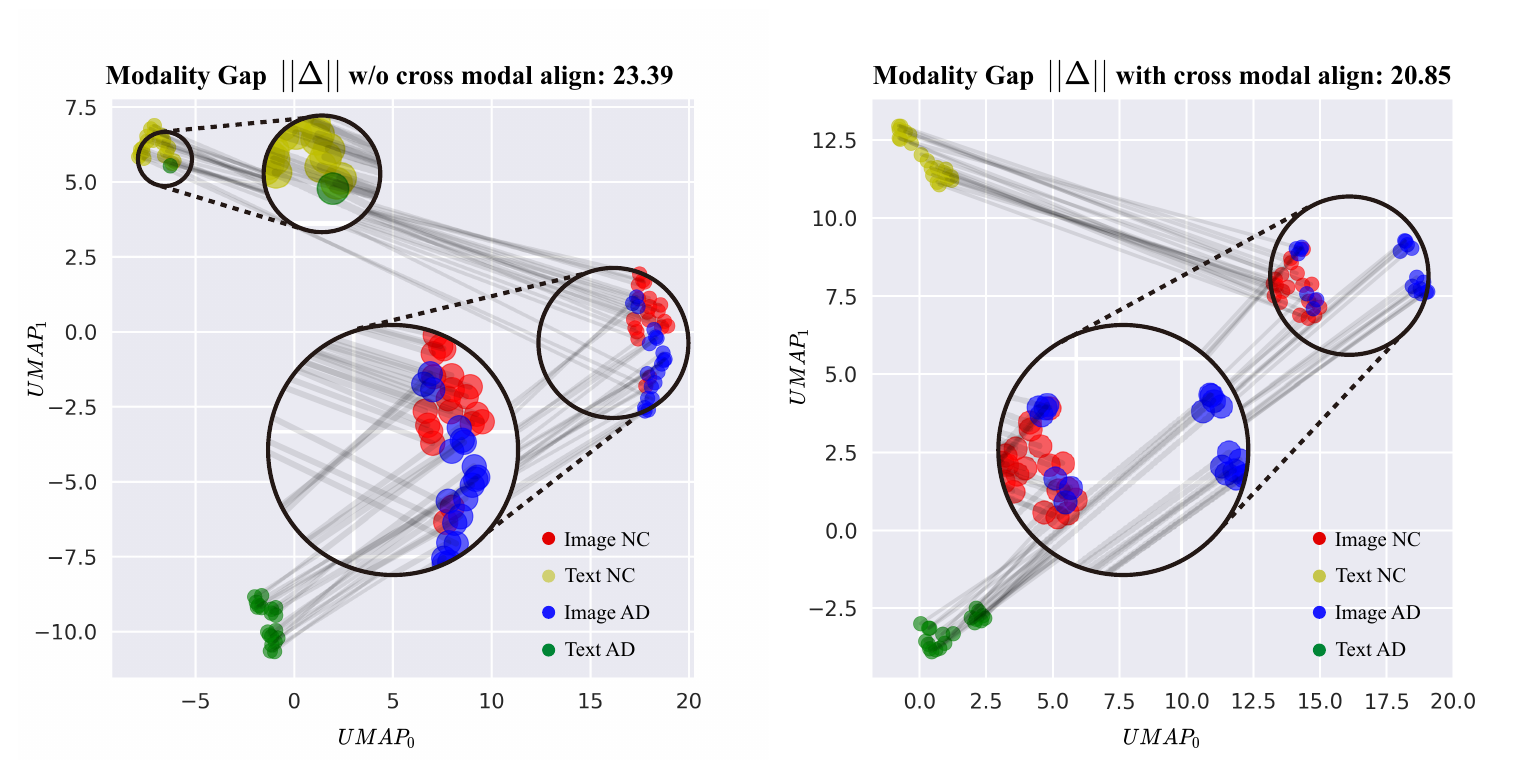}
  \caption{The UMAP visualization of generated image and text embeddings of AD and NC subjects from the ADNI tests set. Black lines: image-text pairs. 
  }
  \label{fig:umap}
\end{figure}

  

\noindent
\subsection{Experimental Results.} 
Table~\ref{tab:overall} reports the experimental results of Alifuse compared to eight baseline methods across five datasets for classifying Alzheimer's disease, including zero-shot classification scenarios. Overall, most multimodal models outperform image-only and text-only models, highlighting the significance of integrating both imaging and textual data in medical diagnosis tasks. Notably, since the text includes useful information closely related to the diagnostic results, the text-only model outperforms the image-only models on most datasets. 

Alifuse significantly outperforms the image-only and text-only models, as well as the other multimodal models across all datasets. For instance, on the ADNI test set, Alifuse achieves the highest accuracy of 87.03\%, over 12\% higher than the best image-only model, M3T, which uses only structural MRIs as input. Compared to the text-only method, Alifuse shows an improvement of over 5\%. When compared to GIT, Alifuse demonstrates an advantage of over 4.5\%. Even against the current SOTA model, Irene, a transformer-based multimodal classification model, Alifuse delivers competitive results, surpassing Irene by over 3.5\%. We attribute this to Alifuse's superior feature alignment technique and greater flexibility in handling missing data, unlike Irene, which fills missing positions with -1 for structural data. Similar performance improvements are observed on the NACC test set and the three zero-shot test sets. 
Although AIBL collects fewer data types, it focuses on AD biomarker variables. Also, AIBL has less missing data; for example, all subjects in AIBL have MMSE scores, compared to only 88\% in ADNI.
This makes AIBL's non-imaging data more information for AD classification, as evidenced by the better performance of text-based methods. 
\subsection{Ablation Studies}

\begin{table}[t]
\scriptsize
\begin{center}
    \caption{Ablation Experiments on ADNI test dataset. 
    }
    \label{tab:ablation}
    \begin{tabular}{lcc}
    \cline{1-3}
        Task & ACC & AUC \\
    \cline{1-3}
        \textbf{(a) Loss Components} \\
    \cline{1-3}
        $\mathcal{L}_{cls}$ & 84.11\% (2.92\%$\downarrow$) & 88.08\% (2.20\%$\downarrow$)  \\
        $\mathcal{L}_{cls}+\mathcal{L}_{cl}$ & 86.51\%  (0.52\%$\downarrow$) & 89.88\% (0.40\%$\downarrow$)   \\
        $\mathcal{L}_{cls}+\mathcal{L}_{cl}+\mathcal{L}_{res}$(ours) & \textbf{87.03\%} & \textbf{90.28\%}  \\
    \cline{1-3}
        \textbf{(b) Modalities} \\
    \cline{1-3}
        Alifuse-I & 76.85\% (10.18\%$\downarrow$) & 82.63\% (7.65\%$\downarrow$) \\
        Alifuse-T & 83.72\% (3.31\%$\downarrow$) & 87.54\% (2.64\%$\downarrow$) \\
        Alifuse(ours) & \textbf{87.03\%} & \textbf{90.28\%}  \\
    \cline{1-3}
        \textbf{(c) Non-image Data Components} \\
    \cline{1-3}
        w/o Dr.Sum. & 85.80\% (1.23\%$\downarrow$) & 89.35\% (0.93\%$\downarrow$) \\
        w/o LabRes. & 82.14\% (4.89\%$\downarrow$) & 86.61\% (3.67\%$\downarrow$)   \\
        w/o Demogr.  & 83.56\% (3.47\%$\downarrow$) & 87.17\% (3.11\%$\downarrow$) \\
        Dr.Sum.+LabRes.+Demogr.(ours) & \textbf{87.03\%} & \textbf{90.28\%} \\
    \cline{1-3}
    \end{tabular}
  \end{center}
\end{table}

\subsubsection{Impact of loss terms in Alifuse}
We conduct an ablation study to demonstrate the contribution of each loss function in Alifuse, i.e., the classification loss $\mathcal{L}_{cls}$, contrastive loss $\mathcal{L}_{cl}$, and reconstruction loss $\mathcal{L}_{cls}$. The best results are obtained when all three components are included, as reported in Table~\ref{tab:ablation}(a). With only $\mathcal{L}_{cls}$ used, the model performance has reductions of 2.92\% and 2.20\% in accuracy (ACC) and AUC from the best-performing configuration, respectively. Incorporating $\mathcal{L}_{cl}$ alongside $\mathcal{L}_{cls}$ improves performance by 2.5\% in ACC and 1.8\% in AUC. This study establishes the configuration of all losses as the optimal setup for maximizing both metrics.




\subsubsection{Contributions of each modality}

From Table~\ref{tab:ablation}(b), the ablation study demonstrates that the multi-modal design outperforms the unimodal settings, highlighting the necessity of utilizing both imaging and non-imaging data for AD diagnosis. This study also shows that the non-imaging data is more effective in classifying AD compared to imaging data alone, as evidenced by the significant improvement observed when integrating non-imaging text data with imaging data. It suggests that non-imaging data provides critical context and additional diagnostic cues that enhance the overall model performance.

\subsubsection{Contributions of non-imaging data}

We further investigate which type of non-imaging data contributes the most, as reported in Table~\ref{tab:ablation}(c). The classification accuracy and AUC experience the greatest reduction when lab results are removed, which is expected since lab results such as MMSE, CDR, and logical memory are critical for AD diagnosis. Additionally, demographics and the doctor's narrative summary also significantly contribute to improving model performance. 


\subsection{Visualizations}

We visualize the attention areas of our Alifuse using current transformer interpretability techniques~\cite{chefer2021transformer}.
Figure~\ref{fig:attention} displays AD-related activation maps in 3D MRI scans and clinical texts from each dataset. 
The image and text attention maps are generated from the self-attention layer of the last block of the unimodal encoder, which has the highest similarity score with the $[CLS]$ token of the query. 
Regarding images, Alifuse focuses more on brain structures such as the hippocampus, ventricle, and cerebral cortex, which are crucial for AD diagnosis. In terms of text, Alifuse pays more attention to the values of `age' and `education level'. Compared to the word `male', Alifuse assigns more attention to `female', aligning with the higher prevalence of AD among women. Among the lab results, Alifuse emphasizes `MMSE', `CDR', and `logical memory' over `SES', which is only present in the OASIS2 dataset. 


We further explore the influence of our CMA model by visualizing its effect on multimodel alignment. We randomly sample some image-text pairs of AD and NC subjects from the ADNI dataset and create a UMAP visualization~\cite{becht2019dimensionality} of the extracted embeddings from the image and text encoders. With the assistance of CMA, we observe the shortened distance between image and text samples, indicating the alignment among modalities.
Additionally, we quantitatively compute the modality gap $|| \Delta ||$~\cite{liang2022mind}, which represents the difference between the centers of image embeddings and text embeddings. The modality gap decreases from 23.39 to 20.85 after alignment, indicating that CMA enhances collaboration between image and text modalities.
Additionally, CMA helps differentiate indistinguishable modalities by extracting more discriminative features. For instance, in the right image of Fig.~\ref{fig:umap}, the UMAP features of NC and AD images are more easily separable.

\vspace{-0.2em}

\section{Conclusion}

In this paper, we propose a novel framework, Alifuse, that aligns and fuses multimodal medical data in the form of images and texts. This framework combines contrastive and restorative learning in a unified manner. Given Alifuse’s generalizability, we believe it presents a fundamental step towards developing universal representations for multimodal medical diagnosis.
By training and evaluating Alifuse on multiple AD study datasets, we demonstrate its effectiveness in classifying NC, MCI, and AD, consistently outperforming eight recent baselines across all datasets. Currently, we only consider MRI scans, future work will include more image modalities like PET (Positron Emission Tomography). Additionally, incorporating longitudinal data could further improve the diagnosis accuracy.

\vspace{-0.2em}
\section*{Acknowledgment}

This research work was supported by NSFC 62203303 and Shanghai Municipal Science and Technology Major Project 2021SHZDZX0102.

\bibliographystyle{splncs04}
\bibliography{refs}

\begin{thebibliography}{10}
\providecommand{\url}[1]{\texttt{#1}}
\providecommand{\urlprefix}{URL }
\providecommand{\doi}[1]{https://doi.org/#1}

\bibitem{bao2022vlmo}
Bao, H., Wang, W., Dong, L., Liu, Q., Mohammed, O.K., Aggarwal, K., Som, S., Piao, S., Wei, F.: Vlmo: Unified vision-language pre-training with mixture-of-modality-experts. Advances in Neural Information Processing Systems  \textbf{35},  32897--32912 (2022)

\bibitem{becht2019dimensionality}
Becht, E., McInnes, L., Healy, J., Dutertre, C.A., Kwok, I.W., Ng, L.G., Ginhoux, F., Newell, E.W.: Dimensionality reduction for visualizing single-cell data using umap. Nature biotechnology  \textbf{37}(1),  38--44 (2019)

\bibitem{beekly2007national}
Beekly, D.L., Ramos, E.M., Lee, W.W., Deitrich, W.D., Jacka, M.E., Wu, J., Hubbard, J.L., Koepsell, T.D., Morris, J.C., et~al.: The national alzheimer's coordinating center (nacc) database: the uniform data set. Alzheimer Disease \& Associated Disorders  \textbf{21}(3),  249--258 (2007)

\bibitem{chan2020computer}
Chan, H.P., Hadjiiski, L.M., Samala, R.K.: Computer-aided diagnosis in the era of deep learning. Medical physics  \textbf{47}(5),  e218--e227 (2020)

\bibitem{chefer2021transformer}
Chefer, H., Gur, S., Wolf, L.: Transformer interpretability beyond attention visualization. In: Proceedings of IEEE/CVF CVPR. pp. 782--791 (2021)

\bibitem{chen2023medblip}
Chen, Q., Hu, X., Wang, Z., Hong, Y.: Medblip: Bootstrapping language-image pre-training from 3d medical images and texts. arXiv preprint arXiv:2305.10799  (2023)

\bibitem{chen2019med3d}
Chen, S., Ma, K., Zheng, Y.: Med3d: Transfer learning for 3d medical image analysis. arXiv:1904.00625  (2019)

\bibitem{devlin2018bert}
Devlin, J., Chang, M.W., Lee, K., Toutanova, K.: Bert: Pre-training of deep bidirectional transformers for language understanding. arXiv:1810.04805  (2018)

\bibitem{dosovitskiy2020image}
Dosovitskiy, A., Beyer, L., Kolesnikov, A., Weissenborn, D., Zhai, X., Unterthiner, T., Dehghani, M., et~al.: An image is worth 16x16 words: Transformers for image recognition at scale. arXiv:2010.11929  (2020)

\bibitem{ellis2009australian}
Ellis, K.A., Bush, A.I., Darby, D., De~Fazio, D., Foster, J., Hudson, P., Lautenschlager, N.T., Lenzo, N., et~al.: The australian imaging, biomarkers and lifestyle (aibl) study of aging: methodology and baseline characteristics of 1112 individuals recruited for a longitudinal study of alzheimer's disease. International psychogeriatrics  \textbf{21}(4),  672--687 (2009)

\bibitem{he2022masked}
He, K., Chen, X., Xie, S., Li, Y., Doll{\'a}r, P., Girshick, R.: Masked autoencoders are scalable vision learners. In: Proceedings of IEEE/CVF CVPR. pp. 16000--16009 (2022)

\bibitem{he2016deep}
He, K., Zhang, X., Ren, S., Sun, J.: Deep residual learning for image recognition. In: Proceedings of IEEE/CVF CVPR. pp. 770--778 (2016)

\bibitem{jaegle2021perceiver}
Jaegle, A., Gimeno, F., Brock, A., Vinyals, O., Zisserman, A., Carreira, J.: Perceiver: General perception with iterative attention. In: International conference on machine learning. pp. 4651--4664. PMLR (2021)

\bibitem{jang2022m3t}
Jang, J., Hwang, D.: M3t: three-dimensional medical image classifier using multi-plane and multi-slice transformer. In: Proceedings of IEEE/CVF CVPR. pp. 20718--20729 (2022)

\bibitem{liang2022mind}
Liang, V.W., Zhang, Y., Kwon, Y., Yeung, S., Zou, J.Y.: Mind the gap: Understanding the modality gap in multi-modal contrastive representation learning. NeurIPS  \textbf{35},  17612--17625 (2022)

\bibitem{malone2013miriad}
Malone, I.B., Cash, D., Ridgway, G.R., MacManus, D.G., Ourselin, S., Fox, N.C., Schott, J.M.: Miriad—public release of a multiple time point alzheimer's mr imaging dataset. NeuroImage  \textbf{70},  33--36 (2013)

\bibitem{marcus2007open}
Marcus, D.S., Wang, T.H., Parker, J., Csernansky, J.G., Morris, J.C., Buckner, R.L.: Open access series of imaging studies (oasis): cross-sectional mri data in young, middle aged, nondemented, and demented older adults. Journal of cognitive neuroscience  \textbf{19}(9),  1498--1507 (2007)

\bibitem{petersen2010alzheimer}
Petersen, R.C., Aisen, P.S., Beckett, L.A., Donohue, M.C., Gamst, A.C., Harvey, D.J., et~al.: Alzheimer's disease neuroimaging initiative (adni): clinical characterization. Neurology  \textbf{74}(3),  201--209 (2010)

\bibitem{radford2021learning}
Radford, A., Kim, J.W., Hallacy, C., et~al.: Learning transferable visual models from natural language supervision. In: International conference on machine learning. pp. 8748--8763. PMLR (2021)

\bibitem{wang2022git}
Wang, J., Yang, Z., Hu, X., Li, L., et~al.: Git: A generative image-to-text transformer for vision and language. arXiv:2205.14100  (2022)

\bibitem{wang2022medclip}
Wang, Z., Wu, Z., Agarwal, D., Sun, J.: Medclip: Contrastive learning from unpaired medical images and text. arXiv:2210.10163  (2022)

\bibitem{xu2021simple}
Xu, M., Zhang, Z., Wei, F., Lin, Y., et~al.: A simple baseline for zero-shot semantic segmentation with pre-trained vision-language model. arXiv:2112.14757  (2021)

\bibitem{yan2022clinical}
Yan, B., Pei, M.: Clinical-bert: Vision-language pre-training for radiograph diagnosis and reports generation. In: Proceedings of the AAAI Conference on Artificial Intelligence. vol.~36, pp. 2982--2990 (2022)

\bibitem{zhou2023transformer}
Zhou, H.Y., Yu, Y., Wang, C., Zhang, S., Gao, Y., Pan, J., Shao, J., Lu, G., Zhang, K., Li, W.: A transformer-based representation-learning model with unified processing of multimodal input for clinical diagnostics. Nature Biomedical Engineering pp. 1--13 (2023)

\end{thebibliography}

\end{document}